\pdfoutput=1

\documentclass[11pt]{article}

\usepackage[]{EMNLP2023}

\usepackage{times}
\usepackage{latexsym}

\usepackage[T1]{fontenc}

\usepackage[utf8]{inputenc}

\usepackage{microtype}

\usepackage{inconsolata}

\title{Understanding and Mitigating Classification Errors \\Through Interpretable Token Patterns}

\usepackage{amsmath}
\usepackage{amsthm}

\makeatletter
\newcommand\incircbin
{%
  \mathpalette\@incircbin
}
\newcommand\@incircbin[2]
{%
  \mathbin%
  {%
    \ooalign{\hidewidth$#1#2$\hidewidth\crcr$#1\bigcirc$}%
  }%
}
\DeclareRobustCommand{\oland}{\incircbin{\land}}
\DeclareRobustCommand{\oxor}{\incircbin{\times}}

\newcommand{\OLAND}[1]{$\oland$(\textit{#1})}
\newcommand{\OLOR}[1]{\normalfont$\oxor$(\textit{#1})}

\makeatother

\usepackage{booktabs}
\usepackage{pifont}
\usepackage{xspace}
\definecolor{applegreen}{rgb}{0.55, 0.71, 0.0}
\newcommand{\cmark}{\color{applegreen}{\ding{51}}}
\newcommand{\xmark}{\textcolor{red}{\ding{55}}}
\newcommand{\ourmethod}{\textsc{Premise}\xspace}
\newcommand\blfootnote[1]{%
  \begingroup
  \renewcommand\thefootnote{}\footnote{#1}%
  \addtocounter{footnote}{-1}%
  \endgroup
}

\author{Michael A. Hedderich \textsuperscript{*} \\
  Cornell University \\
  \texttt{mah499@cornell.edu} \\\And
  Jonas Fischer\textsuperscript{*}\\
  Harvard University \\
  \texttt{jfischer@hsph.harvard.edu} \\\AND
  Dietrich Klakow\\
  Saarland University \\
  \texttt{dietrich.klakow@lsv.uni-saarland.de}\\\And
  Jilles Vreeken\\
  CISPA\\
  \texttt{jv@cispa.de}}

\begin{document}
\maketitle
\begin{abstract}
State-of-the-art NLP methods achieve human-like performance on many tasks, but make errors nevertheless. Characterizing these errors in easily interpretable terms gives insight into whether a classifier is prone to making systematic errors, but also gives a way to act and improve the classifier. We propose to \textit{discover} those \textit{patterns of tokens} that distinguish correct and erroneous predictions as to obtain global and interpretable descriptions for arbitrary NLP classifiers. We formulate the problem of finding a succinct and non-redundant set of such patterns in terms of the Minimum Description Length principle. Through an extensive set of experiments, we show that our method, \ourmethod, performs well in practice. Unlike existing solutions, it recovers ground truth, even on highly imbalanced data over large vocabularies. In VQA and NER case studies, we confirm that it gives clear and actionable insight into the systematic errors made by NLP classifiers.
\end{abstract}

\section*{Extended Abstract}

As\blfootnote{*equal contribution} much as `to err is human,' NLP models make errors too. Some of these errors are due to noise that is inherent to the process we want to model, and therewith relatively benign. Systematic errors, on the other hand, e.g. those due to bias or misspecification, are much more serious as these lead to models that are inherently unreliable. If we know under what conditions a model performs poorly, we can actively intervene, e.g., by augmenting the training data or fixing preprocessing issues, and so improve overall reliability and performance. Before we can do so, we first need to know whether a classifier makes systematic errors, and if so, how to characterize them in easily understandable terms. 

Given a dataset with labels that specify which instances were classified correctly or incorrectly, we are interested in finding combinations of features that describe where the classifier's predictions are incorrect. For an NLP task, the input features are words or tokens. If, for example, the combination of words ``\textit{how, many}'' is primarily found in misclassified instances, this can indicate that our classifier struggles with the concept of counting. A toy example is visualized in Figure \ref{fig:toy_nlp_example}.

\begin{figure}
	\centering
	\begin{tabular}{l c}
		\toprule
		\multicolumn{2}{l}{Instances \hspace{2.8cm} Correct Prediction?} \\ \midrule
		\textbf{How many} ducks are in the picture? & \xmark \\
		What are the ducks eating? & \xmark \\
		\textbf{How many} roosters are in the puddle? & \xmark \\
		Do you see ducks in the puddle? & \cmark \\ 
		Are there many ducks playing? & \cmark \\ 
		\bottomrule
	\end{tabular}
	\caption{Toy example with input instances and the label specifying if the classifier predicted correctly. The pattern \OLAND{how, many} correlates with misclassification. The word \textit{ducks} is also a frequent pattern but independent of the label and therefore not of relevance.\label{fig:toy_nlp_example}} 
\vspace{-0.7cm}
\end{figure}

Local explanation methods such as LIME \cite{local/ribeiro2016LIME} describe the decision boundary of each instance. In contrast, we are interested in an efficient way to obtain a \emph{global} and \emph{non-redundant} description of our classifier's issues on the given input data. To this end, we turn to pattern mining. Here, a combination of features is a \textit{pattern}, and we look for the set of patterns that best characterizes on which instances the classifier tends to perform poorly. This can be phrased as the more general problem of label description. For data with binary features, we are interested in the associations between the feature data and the labels. We formulate this problem in terms of the Minimum Description Length (MDL) principle. The MDL principle is a formal but practical instantiation of Kolmogorov Complexity and allows to cast our problem in terms of finding the model---the set of patterns---that best compresses the data without loss, measured by the description length of our model class. 

To capture phenomena of text input, e.g., synonyms, we consider a model class representing a rich pattern language that allows us to express conjunctions, mutual exclusivity, and nested combinations thereof. As the search space is twice exponential and does not exhibit any easy-to-exploit structure, we propose the efficient \ourmethod algorithm to heuristically discover the \emph{premises} under which we see the given predictions. To guide the search further, we also leverage classifier-independent word embeddings. A full technical description of our novel approach can be found in \cite{pmlr-v162-hedderich22a}.

\subsection*{Experiments \& Results}

In extensive experiments on synthetic data with known ground truth, we compare \ourmethod against 
over ten different baseline approaches. Some methods directly fail due to prohibitively large running time, not finishing a single run within 12 hours. The remaining seven approaches perform poorly, one common issue being that they find thousands of patterns even when there exist only a few ground truth patterns. Only \ourmethod is able to provide a non-redundant description in the presence of noise and label imbalance, and is able to
easily scale to vocabulary sizes of thousands of tokens. 

\textbf{Understanding Limitations of VQA Models} Visual Question Answering (VQA) is the popular and challenging task of answering textual questions about a given image. We analyze the misclassification of Visual7W \cite{vqa/Zhu2016Visual7W} and LXMERT \cite{vqa/Tan2019LXMERT}, both architectures that were state-of-the-art at their time but performed far from optimal and thus serve as interesting applications for describing misclassification.

\begin{table}
	\small
	\centering
			\vspace{0pt}
			\begin{tabular}{p{3cm}|l}
				\toprule
				pattern & example from the dataset\\ \midrule
				\textit{UNK} & how are the UNK covered \\
				\OLAND{how, many} & how many elephants are there \\
				$\oland$(\textit{what, }$\oxor$(\textit{color, } & what color is the bench \\
				\hspace{0.31cm} \textit{colors, colour})) & \\
				\OLAND{on, top, of} & what is on the top of the cake\\
				\OLAND{left, to} & what can be seen to the left\\
				\OLAND{on, wall, hanging} & what is hanging on the wall\\
				\OLAND{how, does, look} & how does the woman look\\
				$\oland$(\textit{what, does, }$\oxor$(\textit{say, } & what does the sign say\\
				\textit{like, think, know, want})) & \\
				\bottomrule
			\end{tabular}
    \caption{\textit{Example patterns \ourmethod finds for Visual7W.}}
\end{table}

The patterns found by \ourmethod highlight the advantage of the richer pattern language, allowing to find patterns with related concepts such as \OLAND{what, \OLOR{color, colors, colour}}.
Generally, our discovered patterns highlight different types of wrongly answered questions, including counting questions, identification of objects and their colors, spatial reasoning, and higher reasoning tasks like reading signs.  These indicate realistic problems, with the issue of counting having been identified manually in the past as well~\cite{DBLP:conf/iclr/ZhangHP18}.  We also observe that Visual7W and LXMERT share certain issues, but specific problems, such as identifying colors, do not show for the latter. This could be an indicator that more recent classifiers have improved capabilities regarding these problems.

\ourmethod also discovers patterns that are biased towards correct classification, which can indicate issues with the dataset. For instance, \OLAND{who, took, \OLOR{photo, picture, pic, photos, photograph}}, although a difficult question, is nearly always answered by ''photographer`` in this dataset, and thus easy to learn.
Another problematic question is indicated by the pattern \OLAND{clock, time}, where usually the answer is ''UNK`` --- the unknown word token --- due to the limited vocabulary of Visual7W.
The pattern hence indicates a setting where the VQA classifier undeservedly gets a good score.

\textbf{Mitigating NER Classification Errors} Here, we analyze a setting where a Named Entity Recognition classifier might perform well during development, its performance when deployed ``in the wild'' however is much worse. Understanding the difference is important for being able to improve the classifier. With \ourmethod, we can identify issues related to different preprocessing, differing label conventions, and unlabeled data. To empirically validate that the found patterns affect the classifier's performance, we also fine-tune the classifier on pattern-guided data, subselecting samples that show patterns associated with errors, which improves the overall performance significantly compared to finetuning based on a random subset of samples.
The patterns discovered by \ourmethod provide, hence, \textit{actionable} insights into how a classifier can be improved.

\textbf{Try It Yourself!}
For an in-depth analysis of the experiments, we refer to the full publication \cite{pmlr-v162-hedderich22a}. To make our approach easy to use, we provide the PyPremise Python library\footnote{\url{https://github.com/uds-lsv/PyPremise}} which encapsulates our approach and allows the NLP or ML practitioner to get explanations for their classifier with a few lines of code.
\bibliography{abbreviations,custom,bib-paper}
\bibliographystyle{acl_natbib}

\end{document}